\documentclass[10pt, a4paper]{article}
\usepackage{lrec2022} 
\usepackage{multibib}
\newcites{languageresource}{Language Resources}
\usepackage{graphicx}
\usepackage{tabularx}
\usepackage{soul}
\usepackage{tipa}
\usepackage{titlesec}
\titleformat{\section}{\normalfont\large\bfseries\center}{\thesection.}{1em}{}
\titleformat{\subsection}{\normalfont\SmallTitleFont\bfseries\raggedright}{\thesubsection.}{1em}{}
\titleformat{\subsubsection}{\normalfont\normalsize\bfseries\raggedright}{\thesubsubsection.}{1em}{}
\renewcommand\thesection{\arabic{section}}
\renewcommand\thesubsection{\thesection.\arabic{subsection}}
\renewcommand\thesubsubsection{\thesubsection.\arabic{subsubsection}}

\usepackage{epstopdf}
\usepackage[utf8]{inputenc}

\usepackage{hyperref}
\usepackage{xstring}

\usepackage{color}

\title{Building an Endangered Language Resource in the Classroom: \\Universal Dependencies for Kakataibo}

\name{Roberto Zariquiey$^{\rho}$, {\bf \large Claudia Alvarado$^{\rho}$}, {\bf \large Ximena Echevarria$^{\rho}$}, {\bf \large Luisa Gomez$^{\rho}$}, \\ {\bf \large Rosa Gonzales$^{\rho}$}  {\bf \large, Mariana Illescas$^{\rho}$}  {\bf \large, Sabina Oporto$^{\rho}$,} \\ {\bf \large Frederic Blum$^{\beta}$,} {\bf \large Arturo Oncevay$^{\epsilon}$, } {\bf \large Javier Vera$^{\upsilon}$}}

\address{$^{\rho}$Pontificia Universidad Católica del Perú, Peru \\ 
 $^{\beta}$Humboldt-Universität zu Berlin and Leibniz-Zentrum Allgemeine Sprachwissenschaft \\
$^{\epsilon}$University of Edinburgh, Scotland \\
$^{\upsilon}$Pontificia Universidad Católica de Valparaíso, Chile \\ 
         \{rzariquiey, claudia.alvarado, ximena.echevarria, luisa.gomez, a20175617, m.illescasb, sabina.oporto\}@pucp.edu.pe \\ 
        frederic.blum@hu-berlin.de,  
        a.oncevay@ed.ac.uk,  
        javier.vera@pucv.cl
        }

\abstract{
In this paper, we launch a new Universal Dependencies treebank for an endangered language from Amazonia: Kakataibo, a Panoan language spoken in Peru. We first discuss the collaborative methodology implemented, which proved effective to create a treebank in the context of a Computational Linguistic course for undergraduates. Then, we describe the general details of the treebank and the language-specific considerations implemented for the proposed annotation. We finally conduct some experiments on part-of-speech tagging and syntactic dependency parsing. We focus on monolingual and transfer learning settings, where we study the impact of a Shipibo-Konibo treebank, another Panoan language resource. 
 \\ \newline  }

\begin{document}

\maketitleabstract

\section{Introduction}

Kakataibo is a language that belongs to the Panoan family spoken by around 3,000 native speakers in the Amazonian region of Peru. This paper describes the methodology implemented in the context of a regular undergraduate Computational Linguistics course to create a UD treebank for this language, as a strategy to develop significant learning settings and at the same time contribute to the computarization of this endangered language of Peruvian Amazon. The Kakataibo UD treebank would enhance the future development of an NLP toolkit for this language, since it is the main requirement to train a dependency parser. By taking advantage of the preexistence of a UD treebank for another Panoan language, Shipibo-Konibo \cite{vasquez-etal-2018-toward}, in this paper, we conduct some experiments in both monolingual and transfer learning settings. 

The paper is organized as follows. First, §2 presents some background information on the Kakataibo language. Then, §3 describes the methodology implemented in the classroom. §4 introduces the Kakataibo UD treebank. §5 presents the experimentation conducted on part-of-speech tagging and syntactic dependency parsing in both monolingual and transfer learning settings (using the Shipibo-Konibo as a baseline). Finally, §6 summarises the conclusions of this paper.

\section{The Kakataibo Language}

Kakataibo (cbr) is a Panoan language spoken by approximately 3,000 people in the Peruvian departments of Hu\'anuco and Ucayali. The Kakataibo people live in various communities along the Aguayt\'ia, San Alejandro, Shamboyacu, Sungaroyacu and Pisqui Rivers, where the language remains vital despite different degrees of contact between Kakataibo people and non-indigenous populations.
\newcite{zariquiey2011b} distinguishes four living Kakataibo varieties: the Lower Aguayt\'ia/Shamboyacu, Upper Aguayt\'ia, Sungaroyacu and San Alejandro dialects. Nokam\'an, a variety named and minimally documented by \newcite{tessmann1930}, was a fifth variety, now extinct \cite{zariquiey2013}. Among the living varieties, the most divergent is the San Alejandro one, with the Upper Aguayt\'ia and Sungaroyacu varieties being highly similar to each other, and (to a lesser degree) to the Lower Aguayt\'ia variety, which is the one represented in the treebank featured in this paper. The sentences belong to the first author´s database and were gathered in the field between 2007 and 2011. The Lower Aguayt\'ia dialect, studied in this paper, exhibits the phonological inventory given in Tables \ref{CON} and \ref{VOW}. 

\begin{table}[htbp]

\centering
\setlength{\tabcolsep}{2.5pt}
\resizebox{\linewidth}{!}{%
\begin{tabular}{lcccccc}
\hline 
 & labial & alveolar & palatal & retroflex & velar & glottal\\ 
[0.5ex] \hline 
stop & p & t &  &  & k \textipa{k\super w} & \textipa{P} \\
affricate & & \t{ts} & \t{t\textipa{S}}  \\
fricative & & s & \textipa{S} & \textipa{\:s} \\
nasal & m & n & \textltailn \\
flap & & \textipa{R} \\
glide & \textipa{\|`B} \\ [1ex] 
\hline 
\end{tabular} 
}
\caption{Kakataibo consonant inventory} 
\label{CON}
\end{table} 

\begin{table}[htbp]
\footnotesize
\centering 
\begin{tabular}{l l c r} 
\hline 
 & front & central & back\\ [0.5ex] 
\hline 
high & i & \textipa{1}  &        u\\
mid &        e & & o\\
low &  & a\\ [1ex] 
\hline 
\end{tabular} 
\caption{Kakataibo vowel inventory} 
\label{VOW} 
\end{table} 

In terms of its typological profile, Kakataibo is an agglutinative language with synthetic verbal morphology (i.e., we find single verbal words composed of several morphemes). The language is both head and dependent marking, with a complex system of grammatical relations that combines ergative and tripartite (i.e., intransitive subject vs. transitive subject vs. transitive object) alignments in case marking with an accusative alignment in subject cross-referencing on both verbs and a closed set of second position clitics. Clausal constituent order is pragmatically determined, but there is a tendency towards verb-final clauses. Word order in the noun phrase is not fixed and most nominal modifiers can appear either before or after the nominal head. The language also exhibits a rich switch-reference system and pervasive use of nominalizations in discourse. Kakataibo verbs are inherently transitive or intransitive (with almost no labile verbs). The transitivity of the verb, which can only be altered by the use of valence changing markers, is encoded in various parts of the clause. Kakataibo exhibits a complex tense system with several past tense markers. There is a large set of verb morphemes and enclitics of different sorts encoding evidentiality, modality, mood, and a highly unusual typological category called speech genre in \cite{zariquiey2018}. For a full reference grammar with an abundant discussion of each of these features, the readers are referred to \newcite{zariquiey2018}.

\section{Methodology in the Classroom}

Most NLP courses and textbooks focus only on algorithms and mathematical techniques, with much less attention to data collection and processing, among other more practical problems associated with the implementation of NLP projects \cite{vajjala-2021-teaching}. This is a fundamental issue when the audience of the course does not come from Computer Sciences and/or Engineering, as is often the case in the growing body of language technology techniques applied in humanities research around the world \cite{Hinrichs2019,hiippala-2021-applied}. In this scenario, determining how much mathematics/programming/linguistics should be included in an NLP-focused course is neither a trivial nor an easy question \cite{vajjala-2021-teaching}. Taking into consideration the prototypical linguistics students' background, the most suitable approach would keep mathematics to the bare minimum, focusing on the basic programming elements (lists, dictionaries, if-else) that may provide the students with the necessary skills to deal with linguistic corpora, as well as accomplish basic NLP-related tasks \cite{vajjala-2021-teaching}. To produce a significant learning experience, such programming elements must be introduced in the solutions of concrete analytic problems, such as creating and processing real linguistic data. This perspective was taken in the implementation of a Computational Linguistics course for undergraduates, taught by two of the co-authors of this paper at the Humanities Department of the Pontificia Universidad Católica del Perú during the second semester of 2021 (duration: sixteen weeks, four teaching hours per week). The Kakataibo treebank launched in this contribution was one of the research outcomes of this course.

\subsection{Background: Course goals and students}
The course was designed for advanced undergraduate students with extensive knowledge in Linguistics (e.g., phonetics, phonology, morphology and syntax), as well as with training on the grammar of a sample Peruvian languages. It had no prerequisites as it is an optional course. Six linguistics undergraduate students, co-authors of this paper, were enrolled in the course. The class size was small due to the novelty of the course and the reduced Linguistics alumni at the university. The students had a minimal technology background, including some experience in web development, though most of the class had neither a deep knowledge of NLP resources nor a background in programming. Since all were undergraduate students, the materials provided were specially designed to cover topics that address the possible uses of specific NLP resources for conducting linguistic work on Amazonian languages. In line with \newcite{Bender-GEAF07}, the pedagogical goals of the course were:

\begin{itemize}
    \item to give students an introduction to Python programming;
    \item to provide hand-on experience in the analysis of linguistic data; and
    \item to explore the consequences of NLP systems and Computational Linguistic analysis, especially in minority languages. 
\end{itemize}

\subsection{Course content}
This one-semester course was divided into three parts. In the first unit (five weeks, twenty teaching hours), the students participated on hand-on classes to increasing complex Python programming tasks. With this knowledge, in the second unit (six weeks, twenty-four teaching hours), the students analysed linguistic data of several sources: text documents (with the main goal to learn the basics of token/type frequency counts), \textit{UniMorph} annotations \cite{10.1007/978-3-319-23980-4_5} (with the main purpose to learn about Python dictionaries) and structured data from typological databases (like SAILS \cite{sails}) (in order to practice with \textit{csv} files and dictionaries). The third unit of the course (five weeks, twenty teaching hours) was focused on building a new Universal Dependencies \cite{10.1162/coli_a_00402} treebank of a Peruvian minority language: Kakataibo.

\subsection{Collaborative Methodology for the Development of the Language Resource}

To accomplish the annotation task proposed in the third unit of the course, we developed a collaborative methodology and an annotation ecosystem that proved satisfactory to complete our project in a short time and with high-quality outcomes. The principle behind our methodology was to promote a bridge between the growing body of descriptive work on Peruvian languages and NLP initiatives. Although there is still a lot to be done in this respect, during the last 20 years a significant number of detailed typology-oriented reference grammars of Peruvian (and South American) languages have been published \cite{zariquiey_etal_2019}. Such grammars often feature hundreds of fully annotated and parsed example sentences, accompanied by an analytical discussion that provides a sound basis for their interpretation. Therefore, developing treebanks based on these examples does not require advanced knowledge of the grammar of the language, but just a proper understanding of the examples to transform typology-oriented annotations into UD annotations. 

We chose Kakataibo, a Panoan language spoken in Peru, since there was an available full reference grammar of the language \cite{zariquiey2018}, written by the first author of this paper, who was also engaged in the regular teaching of the course. The grammar contains 1012 linguistic examples, three fully annotated complete narratives and a small dictionary. We expect to incorporate a larger set of fully glossed sentences into an extended version of the Kakataibo UD treebank lauched here.

\begin{figure}[ht]
\begin{center}
\includegraphics[scale=0.22]{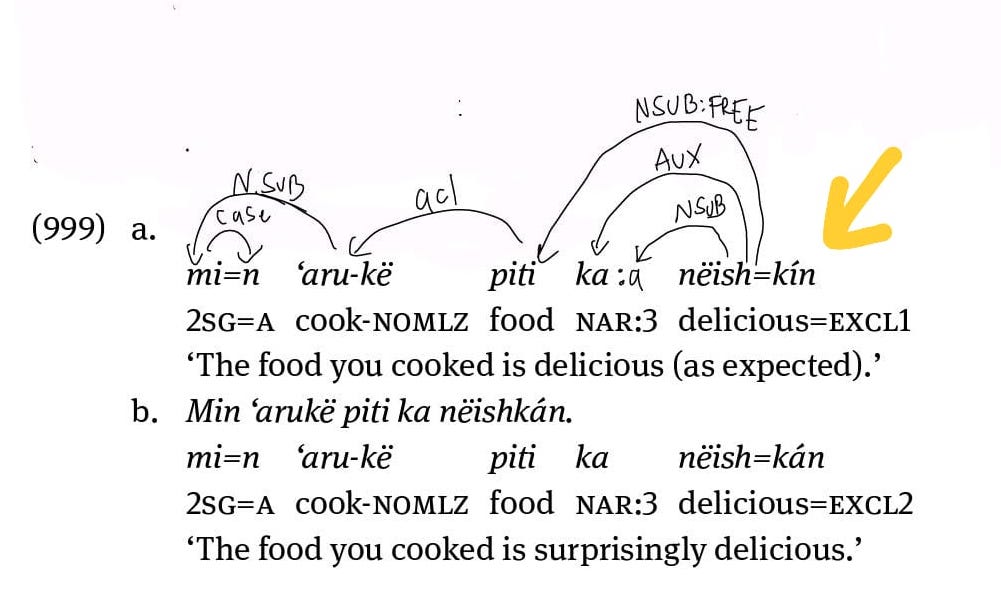} 
\caption{Annotation of a Kakataibo sentence in Zariquiey's grammar}
\label{fig:ex}
\end{center}
\end{figure}

During the third unit of the course, each class consisted of discussions about UD annotations of some illustrative sentences. This served as guidance for the students regarding the parameters of the annotation, to ensure consistency across the annotated sentences. In addition, each student had a small group of test sentences that were reviewed by the expert on Kakataibo (the first author of this paper) to verify the quality of the work. The annotation quality was understood from two points of view: 1° the UD general guidelines; and 2° the particularities of the Kakataibo grammar. Then, each student randomly chooses a set of sentences to annotate and when the complete list of sentences was defined, they were manually annotated by the first author and the students. After that, each student annotated around 15 sentences and problematic cases would be discussed and resolved through Zoom meetings. Figure \ref{fig:ex} features a caption of a manually annotated sentence, the yellow line was added by a student, and indicates that she identified a missing dependency. Once all the corpus was gathered in GitHub, a couple of students, with a supervisor professor, would oversee and correct any possible typing errors and make sure everything was in order. 

Draft annotations were implemented in the annotation tool \textit{UD Annotatrix} \cite{tyers-etal:2018}, and each \textit{CoNLL-U} file was also carefully revised to fix bugs and other technical issues. We ended up with 130 fully annotated sentences from \newcite{zariquiey2018}, which constitute the first Kakataibo UD Treebank. Using the \textit{conllu} Python library \cite{conllu}, CoNLL-U formatted strings were transformed into Python dictionaries to conduct further in-class programming experiments. This process continued for four more weeks after the end of the academic semester, so the total time frame used in the creation of the treebank was nine weeks.

\subsection{Discussion}
An important lesson of this process was the enormous value that grammar's examples have for producing NLP resources. During the treebank creation, it became clear that implementing UD annotation based on typological categories like the ones used in reference grammars is a fairly straightforward process, due to the salient coincidences between linguistic typology and Universal Dependencies \cite{croft2017linguistic}. We strongly believe that the implemented methodology can be efficiently replicated in the future collaborative creation of treebanks based on grammars' examples in the frame of NLP, programming or computational linguistics courses, or workshops for Linguistics students. 

Concerning this methodology's scalability, we propose to include inter-annotator agreement in the following way: bigger classrooms could be divided into groups, each of which would be assigned a set of sentences to be annotated. This dynamic would allow a finer annotation for each sentence. The selected sentences would have to be first analysed by each group member and later on discussed within the group. The final product would show a consensual annotation for that sentence, providing us with a more rigorous filter. In addition to this, we strongly believe that it would be important to maintain the establishment of parameters as a mandatory part of the course. It would help to ensure consistency across the corpus and to reduce the inquiries made to the language expert. Although we benefited from the first author's direct field experience on the language under study, this is not a requisite to successfully implementing this methodology. High-quality reference grammars are often self-explanatory if one has training in linguistic typology.

We believe that increasing the involvement of large groups of undergraduate students would not be an issue as they are generally keen to contribute and get engaged in research projects, particularly those which may contribute to the development and revitalization of endangered languages. The idea of submitting our joint work to an international academic conference was also highly appealing to the students and reinforced their commitment to the project.  

\section{The Kakataibo Treebank}

\begin{figure*}[t!]
\begin{center}
\includegraphics[trim={0.2cm 4.8cm 0.75cm 6cm 0},clip,width=\linewidth]{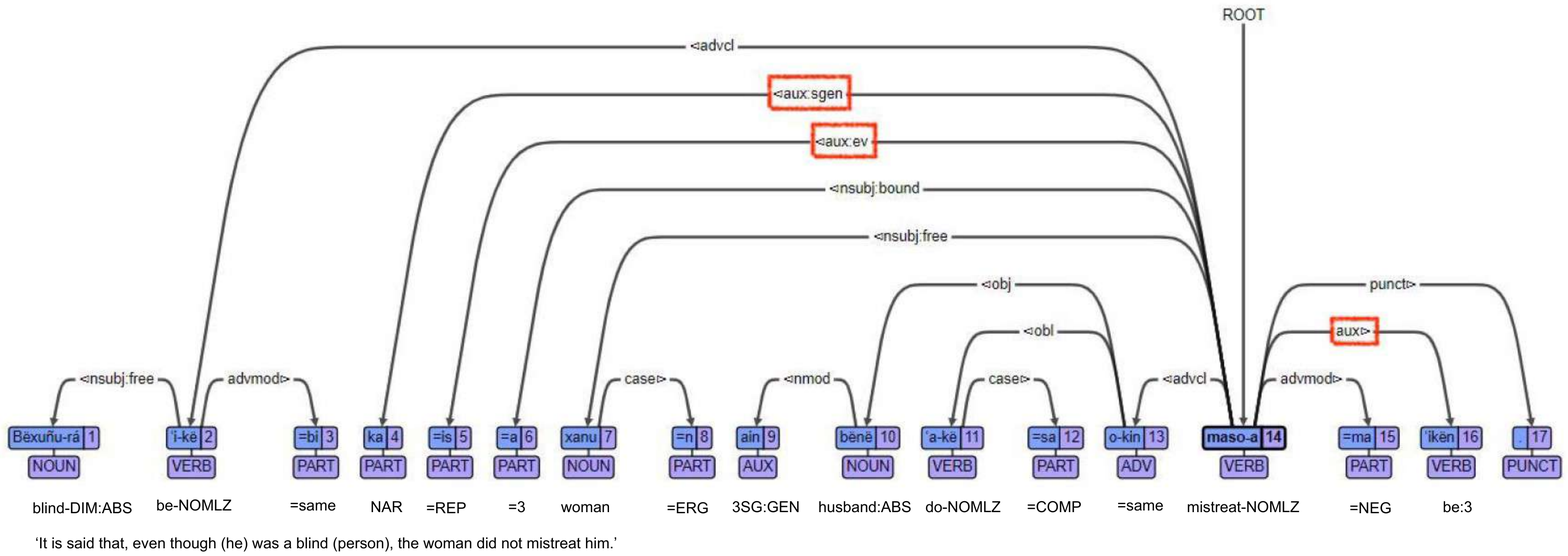}
\caption{A Kakataibo sentence featuring the dependencies \textit{aux:sgen}, \textit{aux:ev} and \textit{aux}. The featured sentence is \textit{Bëxuñurá 'ikëbi kaisa xanun ain bënë 'akësa okin masoama 'ikën} `It is said that, even though her husband was blind, the woman did not mistreat him.’}
\label{fig:aux}
\end{center}
\end{figure*}

\subsection{Part-of-Speech}

Universal Dependencies features a tagset of 17 Part of speech (POS) tags, mainly based on the Google universal part-of-speech tags \cite{petrov-etal-2012-universal}, and 15 of them were used to elaborate the Kakataibo treebank. The POS tags and frequencies in the treebank are shown in Table \ref{POS_list}. The POS tags \textit{X} and \textit{SYM} were not included in this version of the treebank but were relevant for Shipibo-Konibo, another Panoan language \cite{vasquez-etal-2018-toward,pereira-etal-2017-ship}. In the Shipibo-Konibo treebank, \textit{X} was used for onomatopoeias, which is also a part of speech in Kakataibo, not yet attested in the annotated treebank. A future version of this treebank may incorporate those tags in its POS tags repertoire. 

Following the Shipibo-Konibo treebank \cite{vasquez-etal-2018-toward}, Kakataibo enclitics are treated here as an independent closed POS, labelled as \textit{PART}, which is one of the POS tags included in the UD POS tagset. According to \newcite{zariquiey2018}, there are three types of enclitics in Kakataibo: noun phrase (NP) enclitics, second position enclitics, and adverbial enclitics. Noun phrase enclitics appear at the right edge of NPs. Since some NP modifiers are post-nuclear, NP-enclitics do not necessarily attach to nouns, but also adjectives, nominalizations, determinants, and numerals, among other NP modifiers. Second position enclitics are positionally fixed: they always appear after the first constituent of the sentence independently of its syntactic nature. Adverbial enclitics are non-positional and they may appear attached to any constituent of the clause independently of its position. All enclitics in Kakataibo are phrase-level modifiers. 

As can be seen in Table \ref{POS_list}, PART is largely the most frequent POS in the Kakataibo treebank (freq = 0.40), this is mainly because second-position enclitics in Kakataibo are even more obligatory than verbs (there are verbless copula clauses in Kakataibo, but each independent sentence in the language must carry second position enclitics encoding speech genre and subject indexation).

\subsection{Universal Dependency relations}

UD defines a set of 37 dependency relations, mainly based on the Universal Stanford Dependencies \cite{de-marneffe-etal-2014-universal}, and 27 of these have been used for the annotation of the Kakataibo treebank, as specified in Table \ref{dependency_relations_list}. Although UD aims to provide a universal set of syntactic dependencies as a strategy to facilitate consistent annotation across languages and cross-linguistic comparisons \cite{nivre-etal-2020-universal}, it also provides alternatives to code language-specific categories, using "subtype" relation labels. In the case of Kakataibo, it is required to acknowledge the distinction between auxiliary verbs and auxiliary particles. We coded this distinction employing various subtypes of the dependency \textit{aux}, as discussed in \ref{subsec:aux}. On the other hand, in addition to verbal morphology, subject encoding is accomplished through two independent constituents in the Kakataibo sentence: as part of the second-position enclitic complex and utilizing an independent noun phrase (only the former is obligatory). Based on this, we propose two subtypes for the \textit{nsubj} dependency: \textit{bound} and \textit{free}. 

\subsubsection{Subtypes of \textit{aux}}
\label{subsec:aux}

Second position enclitics clearly satisfy the definition of auxiliary provided in the UD protocol, but due to their different POS and their syntactic particularities, they need to be distinguished from verb auxiliaries in periphrastic verbal constructions. We implement this distinction by means of using \textit{aux} without further subtype specification for auxiliary verbs and \textit{aux:subtype} for the various types of categories encoded by means of second-position enclitics. Thus, we have the following \textit{aux} subtypes: \textit{aux:sgen} (related to the obligatory category of speech genre coded in each Kakataibo sentence, which encodes a pragmatic distinction between narrative and conversational genres); \textit{aux:ev} (used for the reportative evidential); \textit{aux:int} (used for the interrogative enclitic); and \textit{aux:dub} (used for the dubitative enclitic). There are more second-position enclitics in Kakataibo, but they have not appear in our treebank yet. A detailed discussion on the syntax and semantic of these enclitics is offered somewhere else \cite{zariquiey2018}. The Shipibo-Konibo treebank follows a similar approach, by including an \textit{aux} subtype (\textit{aux:val}) \cite{vasquez-etal-2018-toward}, which is more or less equivalent to \textit{aux:ev} in Kakataibo (here we used \textit{aux:ev}, since ``evidential'' is more widely used than ``validator'' in the contemporary typological literature).

One interesting point about the distinction between the different types of \textit{aux} dependencies has to do with the direction, auxiliary verbs appear to the right of the root, whereas second-position enclitics appear to the left of the root. A Kakataibo sentence featuring the dependencies \textit{aux:sgen}, \textit{aux:ev} and \textit{aux} is presented in Figure \ref{fig:aux}.

\subsubsection{Subtypes of \textit{nsubj}}
\label{subsec:free}

\begin{figure}[t!]
\begin{center}
\includegraphics[trim={1.9cm 3.15cm 1.55cm 4.5cm},clip,width=\linewidth]{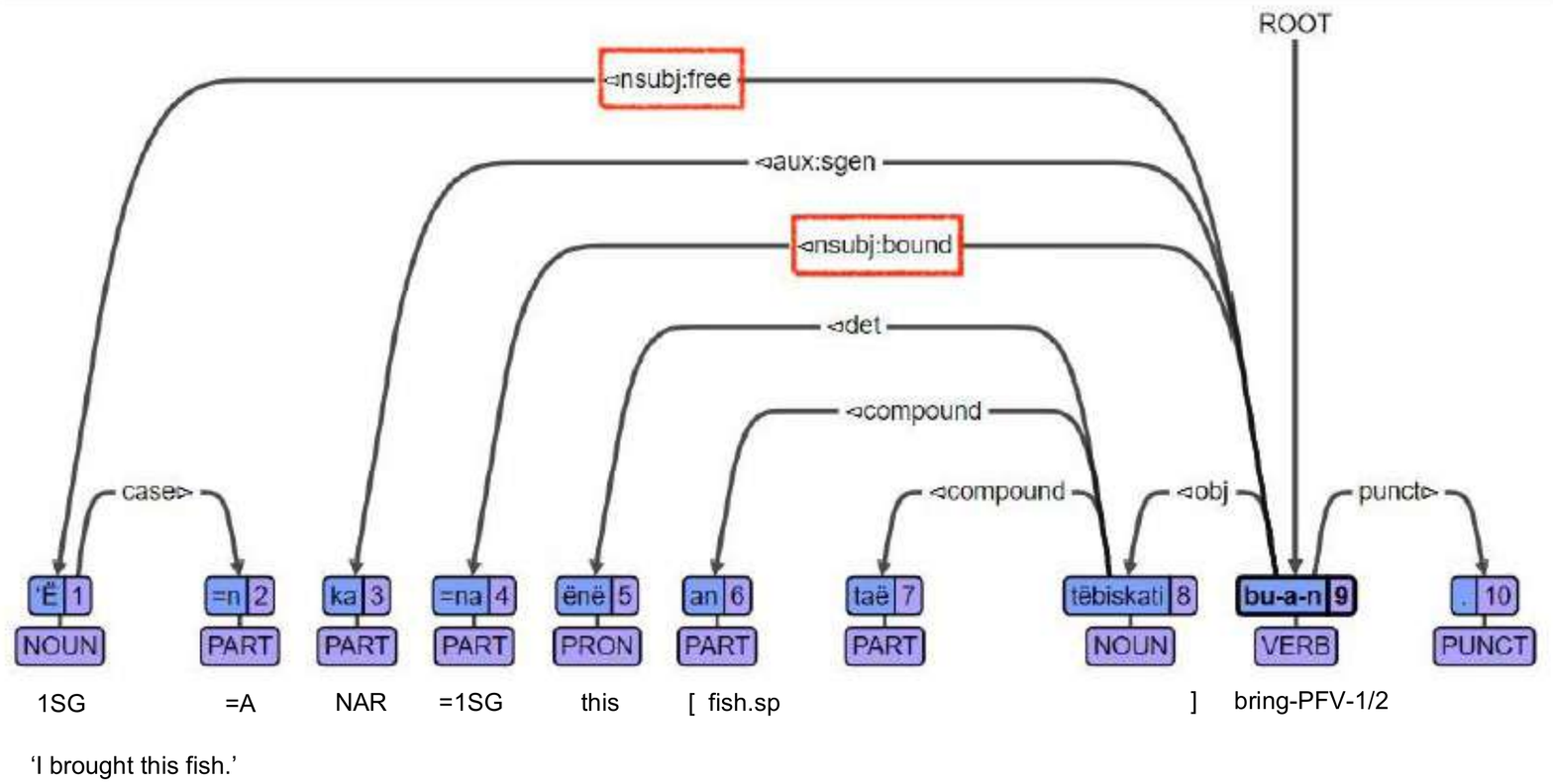} 
\caption{A Kakataibo sentence featuring the dependencies \textit{nsubj:free} and \textit{nsubj:bound}. The featured sentence is \textit{'Ën kana ënë an taë tëbiskati buan} `I took this \textit{an taë tëbisbaki} (fish species)'}
\label{fig:nsubj}
\end{center}
\end{figure}

Noun phrases overtly encoding the subject of a clause are not obligatory in Kakataibo, but there is obligatory subject indexation in the verb and in the second position enclitics. While subject indexation in the verb can be considered as part of the verbal morphology, the decision of treating enclitic as independent POS (PART) lead to an annotation in which the \textit{nsubj} dependency goes from the root to both the second-position enclitic encoding subject indexation and to the head of the subject NP (if overtly expressed). To clarify that there are fundamental differences regarding the syntactic nature of the two ways of encoding subjects in Kakataibo (e.g., one is obligatory and the other is optional), we decided to identify two different subtypes of \textit{nsubj}, that is \textit{nsubj:free} and \textit{nsubj:bound}. A Kakataibo sentence featuring the dependencies \textit{nsubj:free} and \textit{nsubj:bound} is presented in Figure \ref{fig:nsubj}.

\section{Experimentation}

Once we finished the current version of the Kakataibo treebank, we conducted experiments in POS tagging and dependency parsing in different monolingual and transfer settings for both the Shipibo-Konibo and the Kakataibo treebank. For this reason, we first compare the frequency of each tag in both treebanks (see Tables \ref{POS_list} and \ref{dependency_relations_list} in the Appendix) and the utterances length (see \ref{fig:utt_length}). Regarding the POS tags, we observe that both treebanks contain similar proportions, although the Kakataibo treebank presents PART more frequently, and contains less punctuation marks. Besides, the utterances in the Kakataibo treebank tend to be longer than those in the Shipibo-Konibo treebank, with a mean length of 9.52 ($\pm$3.22) for Kakataibo, and 7.08 ($\pm$2.92) for Shipibo. Each language has one utterance that did not fit the limits of the graph, with a respective token length of 28 (Kakataibo) and 23 (Shipibo). These differences pose a limitation for the following transfer learning experiments.

\begin{figure}[ht]
    \centering
    \includegraphics[scale=0.495]{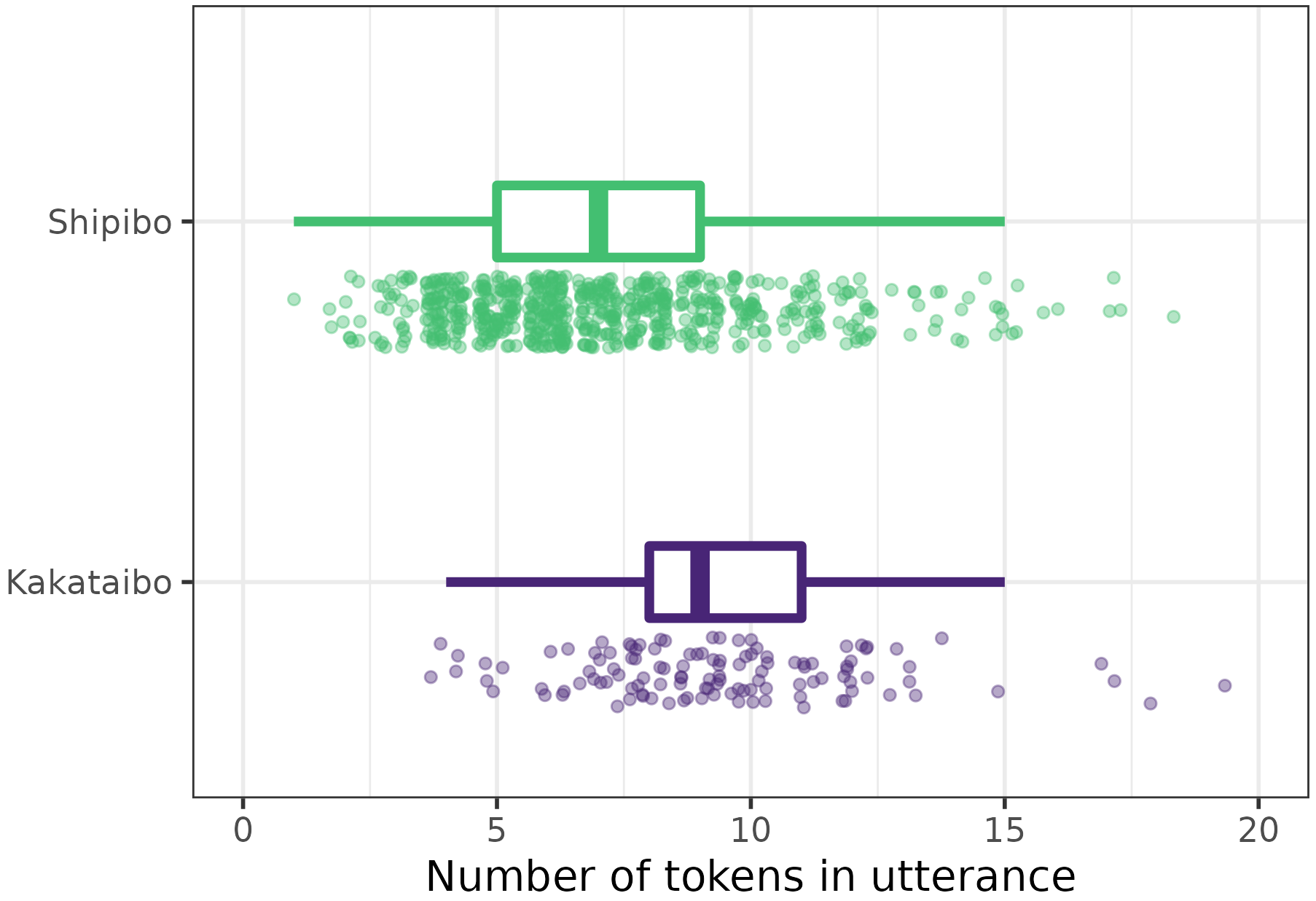}
    \caption{Utterance length in the Panoan treebanks}
    \label{fig:utt_length}
\end{figure}

\begin{table*}[t]
    \centering
    \begin{tabular}{|rll|l|l|l|l|}
      \hline
         & train & fine-tune & cbr accuracy & cbr f1 & shp accuracy & shp f1 \\ \hline\hline
          1 & cbr &  & 84.5±1 & 46.9±2.4 & 35.3±1.2 & 10.6±0.2 \\ 
          2 & shp &  & 61.0±1.4 & 21.1±2.1 & 93.4±0.7 & 84.6±1.6 \\ 
          3 & shp & cbr & 76.9±2.8 & 34.1±4 & 93.2±0.2 & 85.6±0.5 \\ 
       \hline
    \end{tabular}
    \caption{Results of the POS tagging experiment (cbr = Kakataibo, shp = Shipibo-Konibo)}
    \label{table: pos-results}
\end{table*}
\begin{table}[t]
    \centering
    \begin{tabular}{|l|lllr|} \hline
        POS & precision & recall & f1-score & n \\ \hline\hline
        PART   & 0.9681 & 1.0000 & 0.9838 & 91\\
        NOUN   & 0.7317 & 0.8108 & 0.7692 & 37\\
        VERB   & 0.7941 & 0.8710 & 0.8308 & 31\\
        PUNCT  & 1.0000 & 0.9200 & 0.9583 & 25\\
        PRON   & 0.8889 & 0.7273 & 0.8000 & 22\\
        DET    & 0.2500 & 0.2500 & 0.2500 & 4\\
        ADJ    & 0.0000 & 0.0000 & 0.0000 & 3\\
        ADV    & 0.0000 & 0.0000 & 0.0000 & 4\\
        AUX    & 0.6667 & 0.6667 & 0.6667 & 3\\
        PROPN  & 1.0000 & 0.6667 & 0.8000 & 3\\
        NUM    & 0.0000 & 0.0000 & 0.0000 & 2\\
        CCONJ  & 0.0000 & 0.0000 & 0.0000 & 1\\ \hline
        \textbf{micro avg}  & 0.8496 & 0.8496 & 0.8496 & 226\\
        \textbf{macro avg}  & 0.5250 & 0.4927 & 0.5049 & 226\\ \hline
    \end{tabular}
    \caption{f1-scores for POS-tagging}
    \label{table: pos-f1}
\end{table}

\subsection{POS tagging}
For the POS-tagging experiment, Shipibo-Konibo was split into an 80/10/10 training/dev/test set. In order to have sufficient utterances for the dev- and the test-set, Kakataibo was split into partitions of 60/20/20. This ensured that the evaluation was done on more than 20 utterances, even though the training set now only consisted of 72 sentences. While this lead to slightly lower accuracy and f1-scores, it significantly improved the stability of the results and should be considered more reliable\footnote{Another option is to perform a leave-one-out analysis, but given our limited resources, we stick to the partition split.}. The distribution of POS-tags across the sets for Kakataibo are given in Table \ref{POS_split}. We tested three different experiment settings for POS tagging.

\begin{enumerate}
    \item Monolingual training with the Kakataibo treebank.
    \item Monolingual training with the Shipibo-Konibo treebank and zero-shot transfer to Kakataibo.
    \item Monolingual training with the Shipibo-Konibo treebank and fine-tuning for Kakataibo.
\end{enumerate}

As model architecture, we used a BiLSTM-CRF dependency parser implemented in \textit{flair} \cite{akbik2019flair}.\footnote{\href{https://github.com/flairNLP/flair}{https://github.com/flairNLP/flair}, Version 0.10, MIT License} The contextual string embeddings \cite{akbik2018coling} were based on the JW300-corpus \cite{agic-vulic-2019-jw300}, which was specifically trained on typologically diverse low-resource languages, and showed significantly better results than transformer-based embeddings for our experiments. The overall results can be found in Table \ref{table: pos-results} and the f1-scores per POS tag is given in Table \ref{table: pos-f1}.

Despite the small training set, the accuracy in both the fine-tuning setting as well as the monolingual Kakataibo training showed good results. Especially the monolingual training was very successful, with an accuracy over 84\% on average. The low f1-score partially has its origin in the fact that not all tags are equally present in the three different data partitions used for training and testing.

It is also noteworthy that even the second setting, a fully lexicalized zero-transfer of the POS-tagger, achieved an accuracy over 60\%. For a semi-automated workflow of annotating a new treebank, it could prove worthwhile to train a zero-shot model on a closely related language and then correct at least 100 utterances manually. From this point, it would then be recommended to start building a monolingual tagger for further annotations.

\subsection{Dependency parsing}
For dependency parsing, we use a deep bi-affine neural dependency parser \cite{dozat2017deep} that is implemented in supar \cite{zhang-etal-2020-efficient}.\footnote{\href{https://github.com/yzhangcs/parser}{https://github.com/yzhangcs/parser}, Version 1.01, MIT License} The following settings were used in the experiment:

\begin{enumerate}
    \item Delexicalized transfer from Kazakh to (lexical) Kakataibo.
    \item Delexicalized transfer from Shipibo-Konibo to (lexical) Kakataibo.
    \item Delexicalized transfer from Kazakh to delexicalized Kakataibo.
    \item Delexicalized transfer from Shipibo-Konibo to delexicalized Kakataibo.
    \item Monolingual training for Shipibo-Konibo and zero-shot transfer to Kakataibo.
    \item Monolingual training of the reduced Kakataibo set (60/20/20).
    \item Monolingual training with the full Kakataibo set (80/10/10).
\end{enumerate}

\begin{table*}[ht]
    \centering
\begin{tabular}{|rll|l|l|l|l|}
  \hline
 & model & train & UAS cbr & LAS cbr & UAS shp & LAS shp \\ 
  \hline\hline
  1 & delex to lex & ktb & 32.4 & 17.7 & 26.5 & 10.4 \\ 
  2 & delex to lex & shp & 9.5 & 3.1 &  &  \\ 
  3 & delex to delex & ktb & 51.3 & 28.2 & 64.9 & 40.5 \\ 
  4 & delex to delex & shp & 60.4 & 39.3 &  &  \\ 
  5 & mono & shp & 20.3±4.7 & 2±0.5 & 87.7±0.9 & 80.1±1 \\ 
  6 & mono & cbr & 73.1±3.3 & 60.4±3 &  &  \\ 
  7 & mono full & cbr & 77.4±3.7 & 67.4±1.6 &  &  \\ 
   \hline
\end{tabular}
    \caption{Results of the Dependency parsing experiment}
    \label{table: dep-results}
\end{table*}

Experiment 1 and 3 are motivated through previous work on Shipibo-Konibo,   showed that the typological proximity of Kazakh \cite{tyers_tl2015} provides good results for delexicalized transfer of a dependency parser to that language \cite{vasquez-etal-2018-toward}. The goal of this setting is to confirm these results for a second Panoan language, and see whether the results are stable, or only a mix of typological proximity and shared random patterns present in both datasets. The Kazakh data is taken from the current UD release \cite{makazhan_tl2015,tyers_tl2015}.\footnote{\href{https://github.com/UniversalDependencies/UD_Kazakh-KTB/tree/master}{https://github.com/UniversalDependencies/UD\_Kazakh-KTB/tree/master}, Version 2.9, CC BY-SA license.}

\begin{figure}[th]
    \centering
    \includegraphics[scale=0.38]{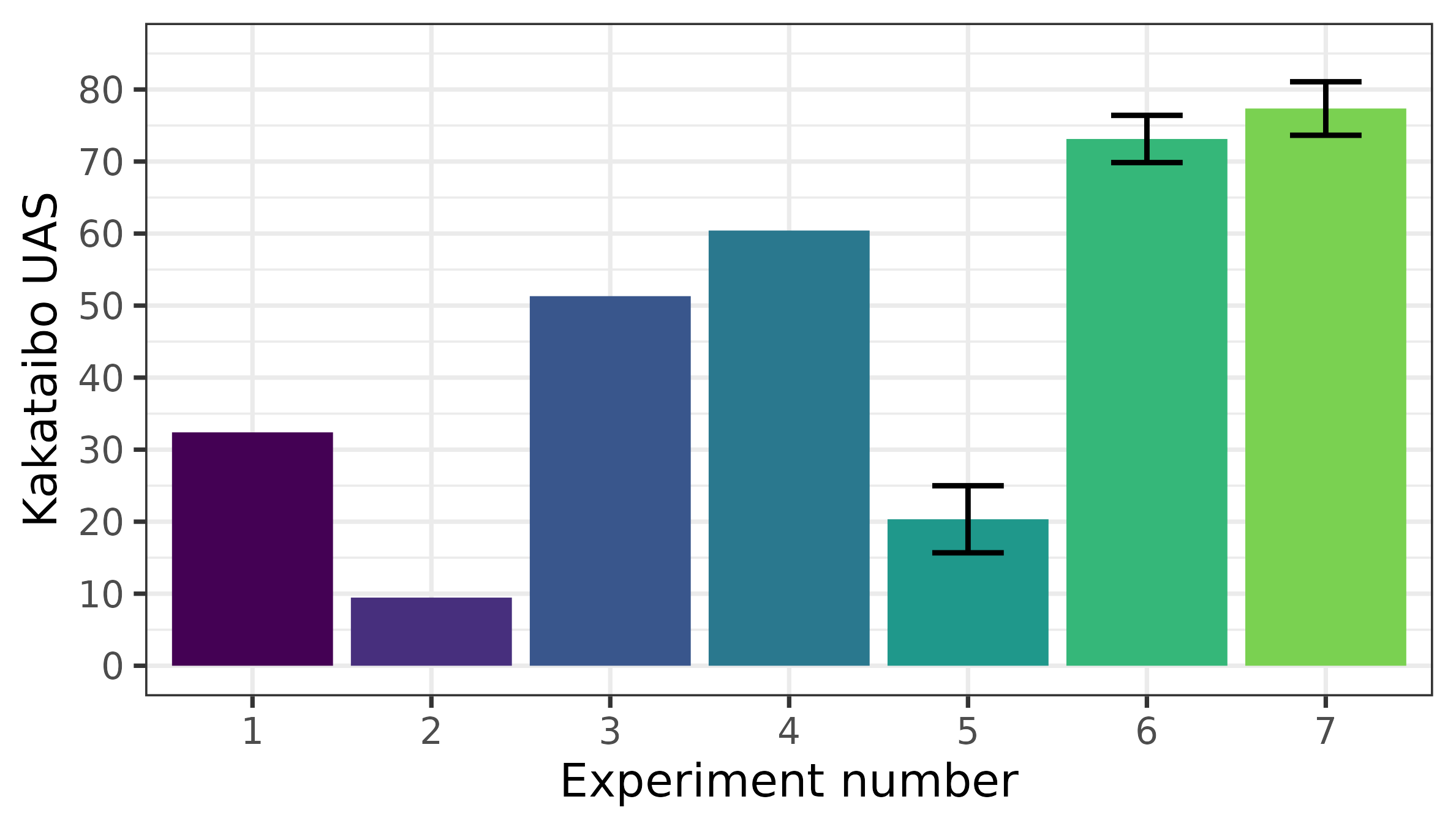}
    \caption{UAS results for Kakataibo}
    \label{fig:dep_results}
\end{figure}

We extend the experiment of delexicalized transfer by adding experiment 2 and 4, and test the delexicalized transfer of two closely-related languages, albeit one, Kakataibo, having less annotated data available \cite{zeman2008cross}. The splits for Experiment 6 and 7 are presented in Table \ref{DEPREL_split}. The results of those experiments are presented in Table \ref{table: dep-results}. The Unlabelled Attachment Score (UAS) refers to the correct assignment of a head for any element, without taking the UPOS into account. The Labelled Attachment Score (LAS) calculates the score for the combination of dependency relation and UPOS. The UAS for Kakataibo in the different experimental settings is presented in Figure \ref{fig:dep_results}.

With respect to experiment 2, it is important to note that the results are not an error. Using the Kazakh embeddings together with the Shipibo-Konibo data yielded results that regularly surpassed 20\% during training, but then collapsed back into <10\%-results for the test data as well. It seems as if the Shipibo-Konibo dependency-relations data is actually surprisingly unfit for transfer to Kakataibo. This holds for the delexical-to-lexical setting, as well as for the lexical zero-shot transfer model. On the other hand, the fully delexicalized results was better than the corresponding Kazakh model. We hypothesise that the origin of this problem lies in the different utterance lengths of the treebanks described earlier.

The findings that delexicalized transfer from Kazakh to Panoan languages can temptatively be confirmed, but not to the same extent as in the previous findings. This suggests that even though the typological proximity has a strong effect on transferability of delexicalized dependency parsers, it may just have been a coincidence that it was Kazakh out of all available typologically similar languages that showed the best results in previous work on Shipibo-Konibo. The exact factors for leveraging typological similarity for sharing NLP resources remain unclear, but further studies on this topic are pressing \cite{bender2009linguistically}.

Comparing experiment 6 and 7, we were again surprised by the small difference between the reduced dataset and the full treebank. This shows that even a small training set of around 100 utterances (74 train + 24 test) shows results that can be implemented in annotation workflows or further experimentation settings. Delexicalized transfer from a closely related language can boost initial annotation steps, once POS tags are already available.

\section{Conclusions}

We introduced here a new NLP resource for a Peruvian endangered language: a Kakataibo Universal Dependencies treebank. The resource comprises 130 annotated sentences, and features 15 POS tags and 27 dependency relations. Two of the Kakataibo dependency relations feature further subtypes: \textit{aux} (\textit{aux, aux:sgen, aux:ev, aux:int} and \textit{aux:dub}) and \textit{nsubj} (\textit{nsubj:free} and \textit{nsubj:bound}. This treebank is the first one produced for Kakataibo, but the second one for a Panoan language, since there is also a UD treebank for Shipibo-Konibo \cite{vasquez-etal-2018-toward}. The existence of a treebank for two Panoan languages allowed us to conduct some experiments on automatic part-of-speech tagging and syntactic dependency parsing in monolingual and transfer learning settings. We did not find a consistently positive impact of transfer learning from the Shipibo-Konibo treebank. However, the results strongly suggest that annotating a small preliminary version of a UD treebank for a minority language can be helpful for reducing annotation efforts in further iterations.  

We also discussed here the collaborative methodology implemented for the creation of the Kakataibo treebank, which was conceived as part a regular Computational Linguistic course for linguistics undergraduates in Peru. The methodology proposed here proved efficient to promote collaborative work among researchers and students in order to produce a full treebank of an endangered language in a limited time frame and in a formative setting. The idea behind the methodology implemented was to promote a bridge between descriptive linguistics and NLP developments, by means of building a UD treebank based on example sentences in published high-quality grammars. We are optimistic about the possibility of replicating our methodology for future similar projects and envisage a near future with larger numbers of endangered languages annotated in the UD framework. 

Finally, the resources and experimentation details for reproducibility are published in: \href{https://github.com/Tarotis/Building-an-Endangered-Language-Resource-in-the-Classroom}{https://github.com/Tarotis/Building-an-Endangered-Language-Resource-in-the-Classroom}

\section*{Acknowledgements}
The first author acknowledges the support of CONCYTEC-ProCiencia, Peru, under the contract 183-2018-FONDECYT-BM-IADT-MU from the funding call E041-2018-01-BM.

\section{Bibliographical References}
\label{reference}
\bibliographystyle{lrec2022-bib}
\bibliography{literature}

\onecolumn
\newpage
\section*{Appendix A: Part-of-speech tags used in the datasets}

\begin{table*}[h]
\centering
\begin{tabular}{|rl|r|r|r|r|}
  \hline
 & upos & n\textsubscript{Shipibo} & n\textsubscript{Kakataibo} & freq\textsubscript{Shipibo} & freq\textsubscript{Kakataibo} \\ 
  \hline\hline
1 & ADJ & 153 &  29 & 0.03 & 0.03 \\ 
  2 & ADP &  36 &   6 & 0.01 & 0.01 \\ 
  3 & ADV & 144 &  30 & 0.03 & 0.03 \\ 
  4 & AUX & 204 &  17 & 0.04 & 0.02 \\ 
  5 & CCONJ &  91 &   4 & 0.02 & 0.00 \\ 
  6 & DET & 133 &  33 & 0.03 & 0.03 \\ 
  7 & INTJ &  35 &   1 & 0.01 & 0.00 \\ 
  8 & NOUN & 646 & 216 & 0.14 & 0.21 \\ 
  9 & NUM &  26 &   5 & 0.01 & 0.00 \\ 
  10 & PART & 956 & 415 & 0.20 & 0.40 \\ 
  11 & PINT &   5 &  & 0.00 &  \\ 
  12 & PRON & 451 &  78 & 0.10 & 0.07 \\ 
  13 & PROPN &  58 &  15 & 0.01 & 0.01 \\ 
  14 & PUNCT & 867 & 127 & 0.19 & 0.12 \\ 
  15 & SCONJ &   1 &   2 & 0.00 & 0.00 \\ 
  16 & SUFN &   2 &  & 0.00 &  \\ 
  17 & SUFV &   5 &  & 0.00 &  \\ 
  18 & SYM &   4 &  & 0.00 &  \\ 
  19 & VERB & 855 & 164 & 0.18 & 0.16 \\ 
  20 & VERB\_AUX &   1 &  & 0.00 &  \\ 
  21 & X &   7 &  & 0.00 &  \\  \hline
  & \textbf{total} & \textbf{4680} & \textbf{1142} & \textbf{1} & \textbf{1} \\ \hline
\end{tabular}
\caption{Part-of-speech tags in the datasets. For both the Shipibo-Konibo and Kakataibo datasets, we report the number of annotated tags (n) and the proportion of each tag with respect to all the tags in the dataset (freq).}
\label{POS_list}
\end{table*}

\begin{table*}[ht]
\centering
\begin{tabular}{|rl|r|r|r|}
  \hline
 & upos & Training & Dev & Test \\ 
  \hline\hline
1 & ADJ &  19 &   7 &   3 \\ 
  2 & ADP &   6 &   0 &   0 \\ 
  3 & ADV &  17 &   9 &   4 \\ 
  4 & AUX &   9 &   5 &   3 \\ 
  5 & CCONJ &   3 &   0 &   1 \\ 
  6 & DET &  15 &  14 &   4 \\ 
  7 & INTJ &   1 &   0 &   0 \\ 
  8 & NOUN & 128 &  51 &  37 \\ 
  9 & NUM &   3 &   0 &   2 \\ 
  10 & PART & 235 &  89 &  91 \\ 
  11 & PRON &  41 &  15 &  22 \\ 
  12 & PROPN &  11 &   1 &   3 \\ 
  13 & PUNCT &  76 &  26 &  25 \\ 
  14 & SCONJ &   2 &   0 &   0 \\ 
  15 & VERB &  97 &  36 &  31 \\ 
   \hline
   & \textbf{total} & \textbf{663} & \textbf{253} & \textbf{253} \\ \hline
\end{tabular}
\caption{60/20/20 Split for the UPOS-tags of the Kakataibo dataset}
\label{POS_split}
\end{table*}

\newpage
\section*{Appendix B: Dependency relations used in the dataset}
\begin{table*}[ht]
\centering
\begin{tabular}{|rl|r|r|r|r|}
  \hline
 & deprel & n\textsubscript{Shipibo} & n\textsubscript{Kakataibo} & freq\textsubscript{Shipibo} & freq\textsubscript{Kakataibo} \\ 
  \hline\hline
1 & acl &   5 &  & 0.00 &  \\ 
  2 & advcl & 114 &  40 & 0.02 & 0.04 \\ 
  3 & advmod & 144 &  48 & 0.03 & 0.05 \\ 
  4 & amod & 130 &  25 & 0.03 & 0.02 \\ 
  5 & appos &   6 &   8 & 0.00 & 0.01 \\ 
  6 & aux & 213 &  14 & 0.05 & 0.01 \\ 
  7 & aux:val & 234 &  & 0.05 &  \\ 
  8 & case & 573 & 131 & 0.12 & 0.13 \\ 
  9 & cc &  92 &   1 & 0.02 & 0.00 \\ 
  10 & ccomp &   1 &   4 & 0.00 & 0.00 \\ 
  11 & compound &  79 &   9 & 0.02 & 0.01 \\ 
  12 & conj &  41 &   2 & 0.01 & 0.00 \\ 
  13 & cop & 137 &   2 & 0.03 & 0.00 \\ 
  14 & det & 139 &  20 & 0.03 & 0.02 \\ 
  15 & discourse &   7 &   5 & 0.00 & 0.00 \\ 
  16 & flat &   9 &   1 & 0.00 & 0.00 \\ 
  17 & iobj &  22 &   2 & 0.00 & 0.00 \\ 
  18 & Lfcl & 183 &  & 0.04 &  \\ 
  19 & marker &   1 &  & 0.00 &  \\ 
  20 & nmod &  69 &  76 & 0.01 & 0.07 \\ 
  21 & nsubj & 538 &  & 0.11 &  \\ 
  22 & nummod &  17 &  21 & 0.00 & 0.02 \\ 
  23 & obj & 256 &  66 & 0.05 & 0.06 \\ 
  24 & obl & 123 &  48 & 0.03 & 0.05 \\ 
  25 & punct & 865 & 127 & 0.18 & 0.12 \\ 
  26 & root & 667 & 120 & 0.14 & 0.11 \\ 
  27 & vocative &   2 &   1 & 0.00 & 0.00 \\ 
  28 & x &   1 &  & 0.00 &  \\ 
  29 & xcomp &  12 &  & 0.00 &  \\ 
  30 & aux:dub &  &   1 &  & 0.00 \\ 
  31 & aux:ev &  &  22 &  & 0.02 \\ 
  32 & aux:int &  &   1 &  & 0.00 \\ 
  33 & aux:sgen &  & 121 &  & 0.12 \\ 
  34 & csubj &  &   1 &  & 0.00 \\ 
  35 & dislocated &  &   1 &  & 0.00 \\ 
  36 & list &  &   2 &  & 0.00 \\ 
  37 & nsubj:bound &  & 116 &  & 0.11 \\ 
  38 & nsubj:free &  &  97 &  & 0.09 \\ 
  39 & parataxis &  &   9 &  & 0.01 \\ \hline
    & \textbf{total} & \textbf{4680} & \textbf{1142} & \textbf{1} & \textbf{1} \\ \hline
\end{tabular}
\caption{Dependency relations used in the datasets. For both the Shipibo-Konibo and Kakataibo datasets, we report the number of annotated tags (n) and the proportion of each tag with respect to all the tags in the dataset (freq).}
\label{dependency_relations_list}
\end{table*}

\begin{table*}[ht]
\centering
\begin{tabular}{|rl|rrr|rrr|}
  \hline
  & & \multicolumn{3}{c|}{\textbf{Experiment 6 (60/20/20)}} & \multicolumn{3}{c|}{\textbf{Experiment 7 (80/10/10)}} \\ \hline
 & \textbf{deprel} & Training & Dev & Test & Training & Dev & Test \\ 
  \hline\hline
1 & advcl &  25 &   8 &   7 &  30 &   5 &   5 \\ 
  2 & advmod &  27 &  13 &   8 &  41 &   2 &   5 \\ 
  3 & amod &  15 &   7 &   3 &  22 &   2 &   1 \\ 
  4 & appos &   6 &   1 &   1 &   8 &   0 &   0 \\ 
  5 & aux &   6 &   5 &   3 &  11 &   2 &   1 \\ 
  6 & aux:dub &   1 &   0 &   0 &   1 &   0 &   0 \\ 
  7 & aux:ev &  10 &   8 &   4 &  15 &   3 &   4 \\ 
  8 & aux:int &   0 &   0 &   1 &   1 &   0 &   0 \\ 
  9 & aux:sgen &  68 &  27 &  26 &  94 &  13 &  14 \\ 
  10 & case &  76 &  27 &  28 & 103 &  16 &  12 \\ 
  11 & cc &   0 &   0 &   1 &   0 &   0 &   1 \\ 
  12 & ccomp &   4 &   0 &   0 &   4 &   0 &   0 \\ 
  13 & compound &   9 &   0 &   0 &   9 &   0 &   0 \\ 
  14 & conj &   1 &   0 &   1 &   1 &   0 &   1 \\ 
  15 & cop &   2 &   0 &   0 &   2 &   0 &   0 \\ 
  16 & csubj &   1 &   0 &   0 &   1 &   0 &   0 \\ 
  17 & det &  10 &   9 &   1 &  13 &   5 &   2 \\ 
  18 & discourse &   3 &   1 &   1 &   5 &   0 &   0 \\ 
  19 & dislocated &   1 &   0 &   0 &   1 &   0 &   0 \\ 
  20 & flat &   1 &   0 &   0 &   1 &   0 &   0 \\ 
  21 & iobj &   1 &   1 &   0 &   2 &   0 &   0 \\ 
  22 & list &   0 &   2 &   0 &   0 &   2 &   0 \\ 
  23 & nmod &  42 &  18 &  16 &  59 &   6 &  11 \\ 
  24 & nsubj:bound &  67 &  25 &  24 &  92 &  13 &  11 \\ 
  25 & nsubj:free &  59 &  17 &  21 &  77 &  11 &   9 \\ 
  26 & nummod &  13 &   4 &   4 &  17 &   2 &   2 \\ 
  27 & obj &  40 &  15 &  11 &  53 &   5 &   8 \\ 
  28 & obl &  23 &  12 &  13 &  35 &   7 &   6 \\ 
  29 & parataxis &   4 &   3 &   2 &   7 &   0 &   2 \\ 
  30 & punct &  76 &  26 &  25 & 101 &  12 &  14 \\ 
  31 & root &  72 &  24 &  24 &  96 &  12 &  12 \\ 
  32 & vocative &   0 &   0 &   1 &   1 &   0 &   0 \\ 
   \hline
 & \textbf{total} & \textbf{663} & \textbf{253} & \textbf{253} & \textbf{903} & \textbf{118} & \textbf{121} \\ \hline
\end{tabular}
\caption{Split for the dependency relations for the Kakataibo dataset}
\label{DEPREL_split}
\end{table*}

\end{document}